\documentclass[letterpaper]{article} 
\usepackage{aaai2027}  
\usepackage[hyphens]{url}  
\usepackage{graphicx} 
\usepackage{natbib}  
\usepackage{caption} 
\usepackage{algorithm}
\usepackage{algorithmic}

\usepackage{newfloat}
\usepackage{listings}
\DeclareCaptionStyle{ruled}{labelfont=normalfont,labelsep=colon,strut=off} 
\floatstyle{ruled}
\newfloat{listing}{tb}{lst}{}
\floatname{listing}{Listing}

\usepackage{booktabs}

\title{Create Your Own Work Stack: Deploying Self-hosted AI and Tooling in a Tech Startup}

\author{
Saad Ahmed Bazaz\equalcontrib\corresponding,
Talha Amjad\equalcontrib,
Abdurrehman Subhani\equalcontrib,
Sohaib Ahmed Bazaz,
Abdullah Aamir
}
\affiliations{
Grayhat Developers (Pvt) Ltd\\
\{bazaz, talha.amjad, subhani, sohaib.bazaz, abdullah.aamir\}@grayhat.studio
}

\begin{document}

\maketitle

\begin{abstract}
Self-hosting AI is often presented as a way to reduce inference costs and increase control, but deployment introduces its own costs. We report our experience deploying a hybrid AI stack at Grayhat Developers, a small software company with an AI-intensive development workflow. Our initial motivation was to counter the growing cost of commercial AI usage: Claude/Codex incurred approximately \$325 per month, with more than 25.93 billion recorded tokens. We experimented with local open-source models as replacements for commercial coding agents, but after learning that their performance and inference speed were insufficient, we adopted a hybrid approach in which commercial agents remained responsible for complex tasks while local models handled lower-complexity tasks, organizational information retrieval, and synchronization with self-hosted tools including Outline, Plane, and ERPNext.

We estimated that approximately 30--40\% of our commercial AI workload was associated with context transfer, documentation synchronization, task management, and related operational coordination. By moving this work to local models, we estimate that approximately 90\% of this coordination workload could be handled without commercial inference, corresponding to an estimated 27--36\% reduction in overall commercial-AI workload or cost-equivalent usage. Our first internal deployment nevertheless failed to achieve adoption because it introduced UX friction. Replacing it with a centrally configured deployment substantially improved feedback and adoption. We conclude that local inference is most useful not as a wholesale replacement for commercial AI, but as part of a hybrid architecture that assigns tasks according to their capability requirements, cost, context demands, and operational sensitivity.

\end{abstract}


\section{Introduction}

Often, the biggest believers are the biggest critics.

We are heavy users of AI at Grayhat. We're a team of under 40 people who build software for startups in San Francisco and develop our own products, and AI-assisted development has become part of our normal workflow. We were therefore not looking for a reason to use less AI. We were looking for a way to use it without allowing the cost and operational complexity to grow with every additional task.

Our immediate problem was commercial AI usage. At one point, five Claude Pro seats could be exhausted around the middle of a billing period. For some routine work, the economics became uncomfortable enough that assigning the work to a human developer was cheaper than continuing to consume commercial model capacity.

What followed was not a clean engineering project. We tried things that did not work, changed direction several times, and discovered that some of our assumptions about AI adoption were wrong --- ultimately documenting a deployment pattern and empirical retrospective showing how a small software organization can use local inference as an organizational-context and orchestration layer while retaining commercial models for capability-intensive development tasks. This can be explained as reminiscent to the Kübler-Ross model of grief \cite{kubler1969death}.

\section{Our Experience}

\subsection{Denial: The Icarus of Vibe-coding}

In early 2026, we took on a series of ambitious hyper-casual game development projects with delivery cycles ranging from one to two weeks per game. The projects ran in parallel and involved fewer people than we would traditionally have assigned to that amount of work. Human involvement was concentrated primarily around reviewing outputs, QA, and final validation. This created a highly AI-dependent workflow.

We tried to keep much of our project context in code and repositories. This worked reasonably well for developers who already understood the project, but it soon disconnected the rest of the organization from that context, including designers and managers.

Our initial hypothesis was that AI would primarily reduce development costs. We instead found friction in collaboration and management, so we connected the AI to MCP servers \cite{mcp2026specification} for each self-hosted tool: Outline \cite{outline} for documentation, Plane \cite{plane} for ticket management, and ERPNext \cite{erpnext} for general management and organizational information. 

The resulting AI usage was already substantial. Seven Claude Team/Claude Code seats cost approximately \$175 per month, while seven Codex subscriptions added approximately \$140; usage-based credits added approximately \$10. Across these services, recorded activity exceeded 25.93 billion tokens at a recurring cost of approximately \$325 per month (Table~\ref{tab:before-local}).

Based on repeated workflow observation, approximately 20\% of a typical usage window was consumed by documentation synchronization and a further 10--15\% by task and project-management synchronization. We therefore estimated that 30--40\% of workload involved context transfer, tool descriptions, MCP calls, and operational synchronization rather than coding or higher-value reasoning. This was an operational estimate rather than a controlled token measurement. Prior work similarly identifies substantial context overhead in MCP-based tool use. \cite{jayanti2026mcp}

\begin{table}[t]

\centering

\caption{AI-assisted development before introducing a local LLM.}

\label{tab:before-local}
\small

\begin{tabular}{l r}

\hline

\textbf{Metric} & \textbf{Value} \\

\hline

Monthly commercial AI cost & \$325 \\

Recorded token usage & 25.93B+ \\

Estimated coordination workload & 30--40\% \\

Quality & Very High \\

\hline

\end{tabular}

\end{table}

\subsection{Anger: Open-source AI and Its Frustrations}

Our first proposed solution was simple: try to cut out the paid models entirely.

We experimented with self-hosted models, including Kimi \cite{kimi2025k15} and DeepSeek \cite{deepseek2024}, using various harnesses and open-source IDEs on real development tasks.

The experience was frustrating for the engineers, who had prior experience with paid cloud models. The models were useful for some tasks, but the gap became obvious when we asked them to perform more complicated coding work. A model could produce reasonable snippets and handle smaller requests while still being unsuitable as a replacement for the coding agents our developers were already using.

The gap was particularly noticeable for repository-level work requiring large context windows and fast inference. On realistic consumer hardware, the models were either too slow or insufficiently capable of understanding existing code, dependencies, and conventions without affecting unrelated parts of the application. \cite{swebench2024,naturalcodebench2024}

\subsection{Bargaining: Deciding on Tasks and Using a Hybrid Approach}

Instead of trying to replace our commercial coding agents, we took a more hybrid approach. The concept of a single, universal AI model for everyday grunt work was flawed - many tasks, similar to the ones in nature, do not require 100\% of brain power at all times. We therefore decided which tasks could be handled by local models and which required the capabilities of commercial coding agents.

\subsubsection{Approach to the hybrid work stack}

We had already deployed Outline, Plane, and ERPNext using Coolify \cite{coolify} on a machine running on our premises. We tied together authentication for these tools using Authentik \cite{authentik}. We then self-hosted the MCP servers that connected our AI systems to internal tools on the same machine. The hardware configuration is provided in Appendix \ref{app:infrastructure}.

We used a separate machine for AI workloads. Its hardware and software configuration is also provided in Appendix \ref{app:infrastructure}.

We started with Gemma 3 2.7B \cite{gemma2025technical}, and later shifted to the latest available Gemma model at the time of deployment. The LLM was deployed as llama.cpp instances managed by GPUStack \cite{gpustack}. The AI chat interface (explained in sections ahead) was connected to the AI inference server through local IP.

\begin{figure}
    \centering
    \includegraphics[width=\columnwidth]{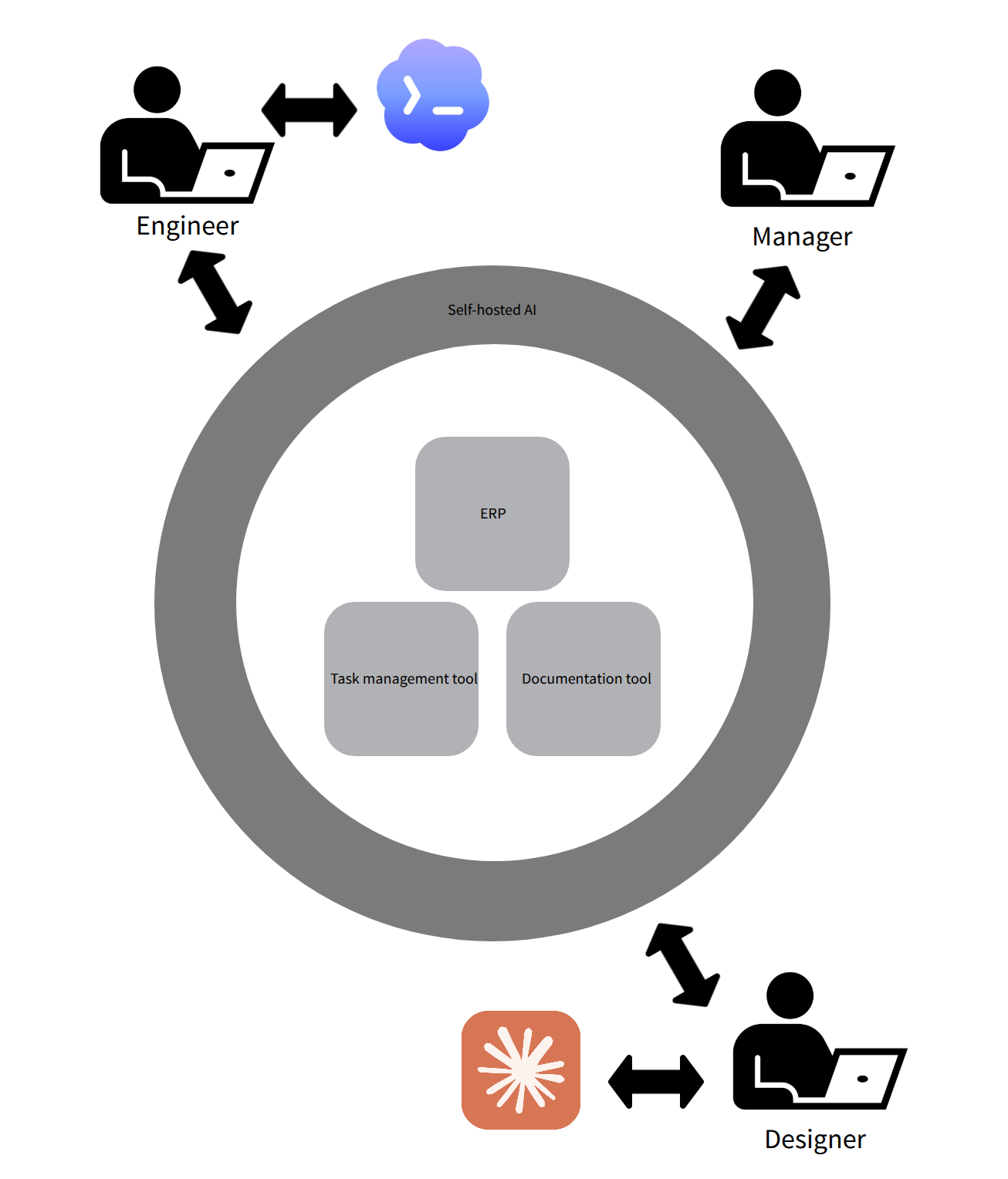}
    \caption{Overview of the hybrid AI workflow process. The self-hosted AI resembles a "moat" around the organization's internal tools.}
    \label{fig:process-ai-moat}
\end{figure}

The process, highlighted in Figure \ref{fig:process-ai-moat}, was roughly as follows:

\begin{enumerate}
\item Humans use local models to retrieve information from Outline and Plane, summarize it, and transform it into a form that can be used by coding agents, such as files in the repository like \texttt{AGENTS.md}.
\item Commercial coding agents remain the preferred option for complex coding tasks and repository-level development. Tasks are written to files in the repository for the coding agents to use.
\item Humans use local models to read the tasks produced by coding agents and translate them into tickets that can be pushed to Plane.
\end{enumerate}

\subsection{Depression: People Still Don't Use It}

Our first internal AI interface, called \textbf{version 1}, was LobeHub \cite{LobeHub}. We set it up on our own devices, and created a tutorial for other users to follow to connect it to our MCP infrastructure. After QA, we released the system for external testing with the rest of the company.

However, the team did not adopt it. The first few demos were great, but people did not actually use it regularly. We initially thought the reason was that people were accustomed to the quality of commercial AI products and therefore found our implementation lower quality. 

It turned out that the biggest hurdle was UX. Despite being a team of engineers, users found it difficult to configure the AI environment reliably. They needed to know which MCP servers were available, how to configure them, and how to authenticate against the systems they wanted to access, especially in a local environment. Commercial AI products had an advantage here because much of this complexity was hidden from the user.

\begin{table}[H]
\centering
\caption{Categorized feedback for version 1 from 11 structured employee interviews.}
\label{tab:gllm-v1-feedback}
\small
\begin{tabular}{l r}
\hline
\textbf{Category} & \textbf{Comments} \\
\hline
Setup \& Onboarding & 8 \\
AI Performance \& Model Quality & 5 \\
Reliability, Uptime \& Guardrails & 2 \\
Tool Integrations & 7 \\
UX \& Trust in Output & 3 \\
Feature Requests & 1 \\
Positive / Neutral Sentiment & 2 \\
\hline
\textbf{Total} & \textbf{28} \\
\hline
\end{tabular}
\end{table}
\subsection{Acceptance: Self-hosted LibreChat and Pre-configured MCPs}

We replaced LobeHub with a self-hosted LibreChat \cite{librechat} deployment, called \textbf{version 2}. With LibreChat, we moved configuration to a central deployment rather than offloading it to users. Users authenticate new MCPs when first accessing the application. The resulting configuration is shown in Appendix \ref{app:frontend}.

This reduced the number of decisions required before use and improved feedback and adoption from the team.

\begin{table}[H]
\centering
\caption{Categorized feedback for version 2 from unstructured user feedback on Discord, with change from version 1.}
\label{tab:gllm-v2-feedback}
\small
\begin{tabular}{l r r}
\hline
\textbf{Category} & \textbf{Comments} & \textbf{Change} \\
\hline
Setup \& Onboarding & 3 & $-62.5\%$ \\
AI Performance \& Model Quality & 6 & $+20.0\%$ \\
Reliability, Uptime \& Guardrails & 7 & $+250.0\%$ \\
Tool Integrations & 6 & $-14.3\%$ \\
UX \& Trust in Output & 5 & $+66.7\%$ \\
Feature Requests & 7 & $+600.0\%$ \\
Positive / Neutral Sentiment & 5 & $+150.0\%$ \\
\hline
\textbf{Total} & \textbf{39} & \textbf{+39.3\%} \\
\hline
\end{tabular}
\end{table}

\section{Learnings}

The original motivation was reducing commercial AI usage and its associated cost. Our observations suggest that approximately 30--40\% of the commercial-AI workload was associated with context synchronization and operational coordination, and approximately 90\% of that workload could be handled by local inference. This corresponds to an estimated displacement of 27--36\% of overall commercial-AI workload, or approximately \$88--117 per month when expressed as a cost-equivalent fraction of the observed \$325 monthly commercial-AI spend (Table \ref{tab:after-local}). The cost still does not drop dead to zero because of equipment costs, electricity cost, and maintenance costs (USD per hour) by the engineers themselves. We view it as moving some costs from recurring SaaS and model usage into infrastructure and engineering overhead, while gaining more control over the systems and the information flowing through them.

Feedback was categorized into seven issue types: setup and onboarding, AI performance and model quality, reliability and guardrails, tool integrations, UX and trust in output, feature requests, and positive or neutral sentiment. Each distinct reported observation was counted once within its corresponding category. Feedback for v1 is visible in Table \ref{tab:gllm-v1-feedback}, while feedback for v2 is visible in Table \ref{tab:gllm-v2-feedback}. The change in feedback between v1 and v2 is also summarized in Table \ref{tab:gllm-v2-feedback}. Because v1 and v2 feedback were collected using different mechanisms and at different stages of deployment, these counts are descriptive rather than a controlled measure of improvement.
\begin{table}[t]

\centering

\caption{Estimated commercial-AI cost-equivalent exposure avoided after introducing a local LLM.}

\label{tab:after-local}
\small
\begin{tabular}{l r}

\hline

\textbf{Metric} & \textbf{Value} \\

\hline

Monthly commercial AI cost baseline & \$325 \\

Estimated coordination workload & 30--40\% \\

Coordination workload handled locally & $\sim$90\% \\

Estimated overall workload displaced & 27--36\% \\

Cost-equivalent workload displaced & $\sim$\$88--\$117/month \\

Local LLM operating cost & \$10--\$30/month \\

Quality & High \\

\hline

\end{tabular}

\end{table}

\subsection{Lessons From the Deployment}

Networking in our internal network was difficult. Setting up routing for LLM inference to a local machine was harder than expected. Server uptime was also a problem --- running the server 24/7 was not needed so we added scheduling, and we added an uninterruptible power supply (UPS) to avoid downtime during power outages.

We also had to add monitoring tools such as Beszel \cite{beszel}, as well as analytics within LibreChat itself, to catch errors early. This meant that, in addition to operational features, we had to build observability as well, and that diverted engineering attention from other commercial work.

Overall, we had to adopt a self-hosting mindset as an entire team. Everyone became responsible for being aware of the servers and for considering what kind of work was being given to each AI model, instead of blindly trusting a commercial model to handle everything. For a small company, this tradeoff needs to be made deliberately.

We also learned that local inference is not always the best option. Research into stronger open-source coding models such as GLM-4 \cite{zeng2024chatglm} continues, but for a small startup, the hardware and setup costs may not be worthwhile for every workload. At the same time, commercial model usage can become expensive when large amounts of context are repeatedly processed. But as mankind and capitalism demand cost-optimization at every stage in tech, higher engineering speed demands that token cost be optimized. So, local inference does not have to replace commercial inference. The two can work in parallel. 

This small example illustrates the distinction: I don't need my model to know Einstein's theory of relativity to write a simple Jira ticket. Assigning simpler tasks to smaller local models can therefore reduce cost without requiring the organization to give up stronger commercial models for tasks where they provide a meaningful advantage. \cite{chen2023frugalgpt,ong2024routellm}

Context is part of the cost. The cost of an AI system is not just the cost of generating an answer. If the model needs to retrieve large amounts of information before it can answer, the retrieval and context-construction process becomes part of the workload.

AI adoption can fail because of UX even when the AI works. Our LobeHub deployment was a useful example. There was no fundamental technical blocker preventing people from using it. The system simply required too much effort from users before it became useful.

The move also reduced our dependence on external AI providers for internal organizational information. Keeping some retrieval and processing local means that not all internal context needs to be sent to an external model provider, hence organizational data can more or less be kept more private than otherwise.

\section{Limitations and Future Work}

Our current setup is still fairly small and manually managed.

The division between local and external models is based largely on our experience with different classes of tasks rather than an automated routing system. We would like to make this more systematic, potentially using a lightweight classifier or policy layer, like Jev \cite{almeida2026jev}, to determine which model should handle a request.

We are also interested in separating model responsibilities further. One possible direction is to use a larger local model primarily for reasoning while using a smaller local model for tool selection and execution. This could reduce the amount of expensive inference required for routine tool interactions.

Another direction is deeper integration with the coding agents our developers already use. At present, the internal AI system and the coding agents are largely separate systems. We would like them to communicate directly, allowing a coding agent to request organizational context from our local agents without requiring a person to manually move that information between systems. However, autonomous behavior might cause another increase in token usage. A query-manager agent, potentially also running locally or at the edge (on the device where the query was originally made), could sit on top of the system and delegate tasks to the coding agent and internal tool agent.

Our cost measurements are operational rather than controlled experiments. We did not establish a formal baseline before beginning the deployment, and several changes happened at approximately the same time. The cost figures therefore describe our experience rather than establishing that self-hosting will produce the same savings for another organization.

We plan to create a replicable environment, provisionally called ``WorkStack,'' as a one-click deployment. This would allow us to reproduce the environment more consistently and measure future changes against a clearer baseline.

\bibliography{aaai2027}


\appendix
\setcounter{secnumdepth}{1}   
\section{Infrastructure Configuration}
\label{app:infrastructure}

\subsection{Self-hosted Tools Machine}

\begin{itemize}
\item Operating system: Ubuntu 24.04.1 LTS
\item Architecture: x86\_64
\item Kernel: 6.8.0-54-generic
\item CPU: 6 cores, Intel Core i5 8th Generation
\item Memory: 16 GB DDR4 RAM
\item Storage: 500 GB SSD
\end{itemize}

\subsection{AI Workload Machine}

\begin{itemize}
\item Operating system: Ubuntu 25.10
\item Architecture: x86\_64
\item Kernel: 6.17.0-35-generic
\item CPU: 20 cores, Intel Core i5 13th Generation
\item GPU: NVIDIA RTX 4080 16 GB
\item Memory: 32 GB DDR5 RAM
\item Storage: 2 TB NVMe SSD
\end{itemize}

\section{Version 2 LibreChat Interface}
\label{app:frontend}

\begin{figure}[H]
    \centering
    \includegraphics[width=\columnwidth]{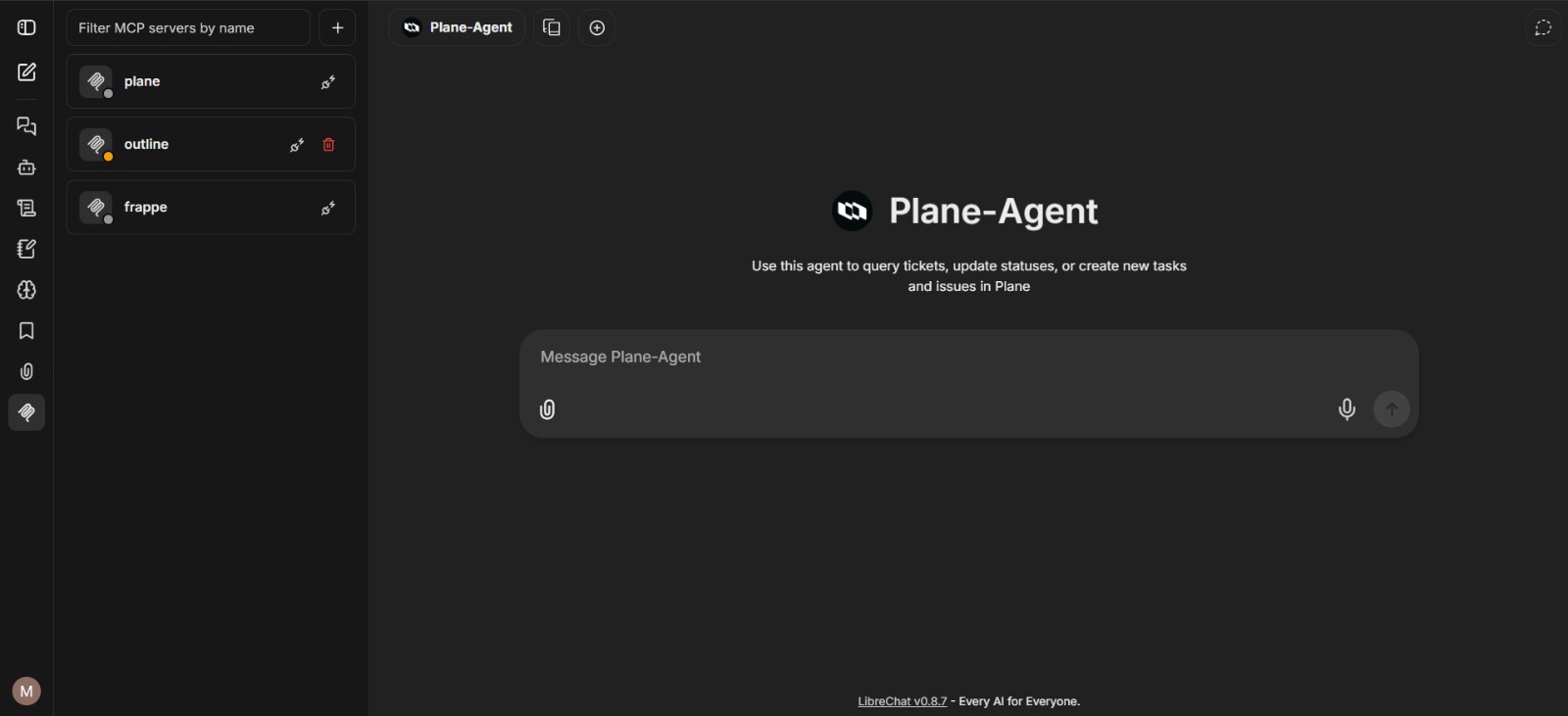}
    \caption{Self-hosted LibreChat with pre-configured MCP connectors and tool-specific agents for Plane, Outline, and ERPNext.}
    \label{fig:librechat-mcp}
\end{figure}

Figure~\ref{fig:librechat-mcp} shows the self-hosted LibreChat deployment with pre-configured MCP connectors, tool-specific agents, and a routing agent.

\end{document}